\documentclass[sigconf]{acmart}
\renewcommand\footnotetextcopyrightpermission[1]{}
\AtBeginDocument{%
  }

\usepackage{graphicx}
\usepackage{booktabs}
\usepackage{algpseudocode}
\usepackage[ruled,vlined]{algorithm2e}
\usepackage{booktabs}
\usepackage{adjustbox}
\usepackage{multirow}
\usepackage{multicol}
\usepackage{tabularx}

\acmSubmissionID{2599}

\begin{document}

\title{Temporal Feature Distillation for Label-Efficient Precise Event Spotting 
in Sports Videos}
\author{Hao Xu}
\affiliation{%
  \institution{Deakin University}
  \city{Melbourne}
  \country{Australia}
}
\email{august.xu@research.deakin.edu.au}

\author{Xinyu Wei}
\affiliation{%
  \institution{Champion Data}
  \city{Melbourne}
  \country{Australia}
}
\email{felix.wei@championdata.com.au}

\author{Sam Wells}
\affiliation{%
  \institution{Paralympics Australia}
  \city{Melbourne}
  \country{Australia}
}
\email{sam.wells@paralympic.org.au}

\author{Sunil Aryal}
\affiliation{%
  \institution{Deakin University}
  \city{Melbourne}
  \country{Australia}
}
\email{sunil.aryal@deakin.edu.au}


\begin{abstract}
Precise Event Spotting (PES) requires distinguishing visually similar yet semantically distinct adjacent frames, making it fundamentally different from image classification and coarse action recognition. Although self-distillation methods such as DINO have shown strong representation learning ability in images, we find that directly applying them to PES is ineffective: without supervised guidance, subtle but crucial motion cues are often suppressed as noise, leading to representations that are insensitive to precise event boundaries. To address this, we propose Temporal Feature Distillation, a semi-supervised objective that aligns temporally informative backbone features, rather than projection-head outputs, to preserve motion-sensitive and boundary-aware cues for frame-level localization. A supervised warm-up with a ramp-up schedule further stabilizes training by ensuring that meaningful event cues are learned before unlabeled distillation begins. We also introduce Transformer Gate Shift, a multi-scale gated shifting module that injects motion-aware temporal information into Vision Transformers. Experiments on four fine-grained sports benchmarks show consistent improvements over fully supervised and semi-supervised baselines. Under 10\% supervision on FSPerf, our method improves mAP by 4.54 points over the strongest competing approach, and with only 80\% labeled data, it matches or surpasses the fully supervised 100\% baseline on two of the four datasets. Code is available at the anonymous repository: \url{https://anonymous.4open.science/r/TFD-8535}.
\end{abstract}




\maketitle
\section{Introduction}
\label{sec:intro}
With the rapid growth of broadcast and amateur recordings, sports videos have become an important source for performance analysis, coaching, and audience engagement \cite{xu2025deeplearningsportsvideo}. Among sports video understanding tasks, detecting the precise moment of key events plays a fundamental role in interpreting game dynamics \cite{hong2022spotting}.

Precise Event Spotting (PES) is a task specifically designed for sports videos, aiming to identify both the temporal location and the category of events—such as a forehand stroke in tennis or a pass in soccer—from long, untrimmed video sequences \cite{hong2022spotting}. Unlike coarse action recognition, PES requires frame-level temporal precision and faces unique challenges inherent to sports scenarios, including rapid motion, frequent occlusion, camera motion, and visually subtle events \cite{hong2022spotting, xarles2023astra, xarles2024t}.

Although recent fully supervised methods have achieved promising results on multiple sports datasets \cite{xu2025multifocustemporalshiftingprecise, hong2022spotting, xarles2024t, tran2024unifying, santra2025precise}, their reliance on dense frame-level annotations limits scalability \cite{schiappa2023self}. Annotating sports videos is time-consuming, often requires expert knowledge, and can be inconsistent when event boundaries are ambiguous. The challenge is even greater in amateur sports videos, where lower frame rates make precise timestamp annotation more difficult.

Another defining characteristic of sports videos is their continuous and large-scale generation, as new matches are recorded daily across professional and amateur levels. Effectively leveraging this ever-growing pool of unlabeled data is therefore essential. In parallel, Unsupervised (USL) and semi-supervised learning (SSL) have gained substantial attention and demonstrated success in various domains, including 2D image classification \cite{caron2021emerging, xie2020self}, segmentation, and general video action recognition \cite{kwon2025boosting, kumar2022end, sohn2020fixmatch, xia2023learning, wang2021self, singh2024semi, xing2023svformer}. Despite this progress, such approaches remain largely underexplored in sports-specific PES, where their potential impact is particularly significant.

\begin{figure*}
    \centering
    \includegraphics[width=1\linewidth]{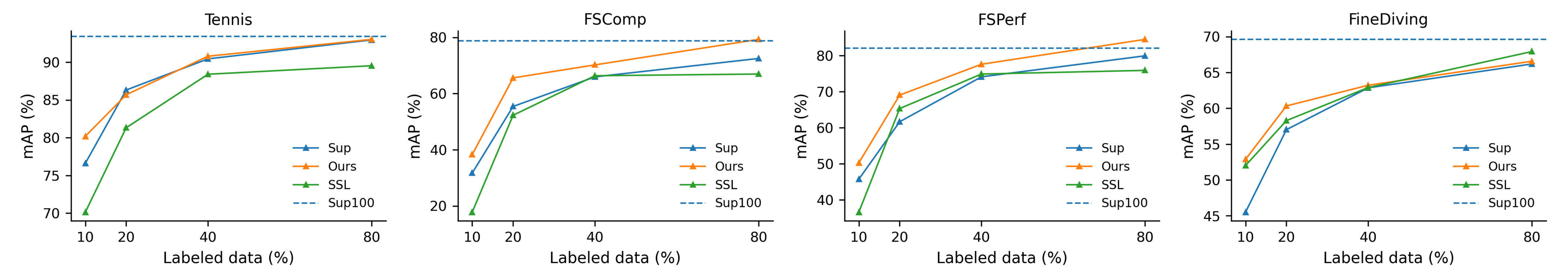}
    \caption{Performance comparison across four sports datasets under varying annotation ratios. Our method consistently outperforms both the supervised baseline and the strongest competing SSL method across label regimes. Using only 80\% labeled data, it matches or surpasses the fully supervised 100\% baseline on 2 of 4 datasets. At 10\% supervision, our method shows consistent gains across all datasets. }
    \label{fig:teaser}
\end{figure*}

Although SSL and USL have been highly effective for image representation learning, directly applying existing self-distillation frameworks to PES is non-trivial. Methods such as DINO \cite{caron2021emerging,oquab2023dinov2,simeoni2025dinov3} align projection-head outputs across augmented views to encourage semantic invariance, which is well suited to image classification. In contrast, PES requires discriminating between temporally adjacent frames that may appear visually similar yet correspond to different event states. Enforcing strong projection-level invariance in this setting can suppress the subtle temporal boundary cues required for precise spotting.

Through systematic analysis (Figure~\ref{fig:cosine_sim}), we observe that DINO-style projection alignment over-smooths representations across neighboring frames, reducing boundary sensitivity that is critical for PES. Motivated by this observation, we propose Temporal Feature Distillation (TFD), which aligns temporally structured backbone features rather than projection-head outputs.

Furthermore, we introduce a temporal-aware Vision Transformer backbone by integrating a multi-scale gated shifting mechanism into a standard 2D ViT. This design enables temporal information exchange across frames while retaining the simple and effective structure of image-based transformers, without relying on dedicated video-specific architectures. In addition, we propose motion-aware and temporally consistent augmentation strategies that explicitly emphasize moving objects—an important characteristic of sports videos, where background appearance often remains largely static and event-relevant information is primarily driven by motion.

Extensive experiments across multiple sports datasets show that our approach consistently outperforms existing SSL methods and strong fully supervised baselines, particularly in low-label regimes. Under only 10\% supervision on FSPerf, it improves mAP by 4.54 points over the strongest competing approach. As shown in Figure~\ref{fig:teaser}, it also matches or exceeds the fully supervised 100\% baseline using only 80\% labeled data on two of the four datasets.

Our key contributions are summarized as follows:
\begin{itemize}
\item \textbf{Temporal Feature Distillation:} a task-driven alternative to DINO-style self-distillation that replaces projection-head alignment with temporally structured backbone-feature alignment, preserving motion-sensitive and boundary-aware cues for frame-level event localization.
\item \textbf{A multi-scale gated shifting module} tailored for Vision Transformers that enables temporal exchange without requiring a dedicated video-transformer design.
\item \textbf{Motion-aware and temporally consistent augmentation strategies} designed specifically for sports videos to prevent trivial learning and enhance temporal continuity.
\end{itemize}

\begin{figure*}
    \centering
    \includegraphics[width=1\linewidth]{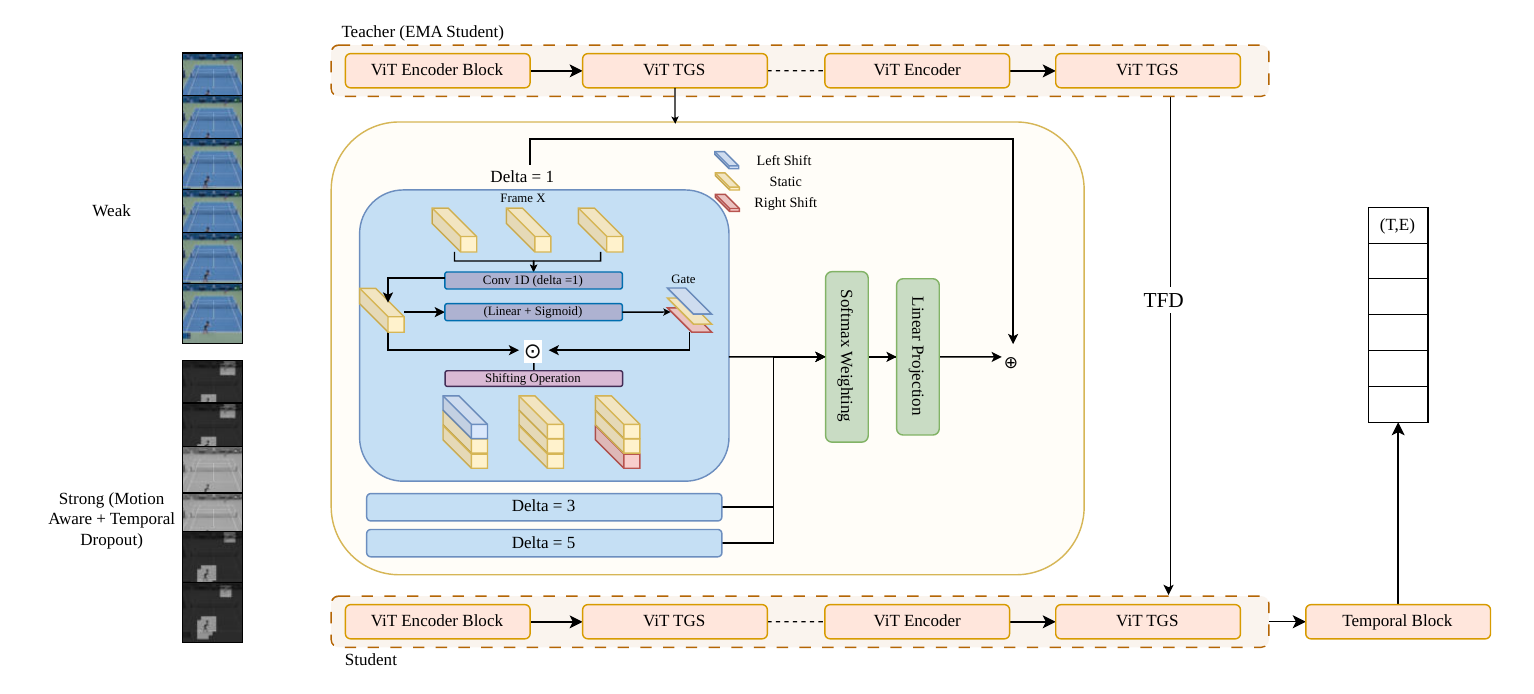}
    \caption{Overall architecture of the proposed SSL framework. The student and EMA teacher networks are built upon ViT encoder blocks with TGS modules and a temporal block. Multi-scale gated temporal shifting and motion-aware augmentations are used to enable effective temporal feature distillation.}
    \label{fig:overview}
\end{figure*}

\section{Related Work}
\subsection{Precise Event Spotting}
PES was originally introduced by Hong et al. \cite{hong2022spotting} as a task specifically designed for sports video understanding. The objective of PES is to predict a set of temporal locations (frame indices) at which events occur, along with their corresponding event classes. A prediction is considered correct if its timestamp falls within a tolerance window of $\delta$ frames around a ground-truth event and the predicted class label is correct, where $\delta$ is typically set between 1 and 3 frames.

Existing PES methods primarily adopt two paradigms for temporal modeling: transformer-based approaches and lightweight shift-based modules. Transformer-based methods \cite{cao2022spotformer, zhu2022transformer, xarles2023astra} and more complex temporal modeling designs \cite{santra2025precise} are capable of capturing long-range temporal dependencies, but often incur substantial computational overhead, typically exceeding 600 GFLOPs. In contrast, lightweight temporal modules such as TSM \cite{lin2019tsm}, GSM \cite{sudhakaran2020gate}, and GSF \cite{sudhakaran2023gate} enable 2D vision backbones to model temporal information efficiently, achieving competitive accuracy with significantly lower computational cost (approximately 40–45 GFLOPs). Owing to this favorable efficiency–accuracy trade-off, shift-based modules have become a core component of several state-of-the-art PES models \cite{hong2022spotting, xarles2024t, tran2024unifying, santra2025precise}.

E2E-Spot \cite{hong2022spotting} established a strong and efficient baseline for PES by integrating GSM into a RegNetY \cite{radosavovic2020designing} backbone and is widely regarded as a foundational model in this domain. Building upon this framework, subsequent works have improved temporal discrimination through finer-grained temporal perception (T-DEED \cite{xarles2024t}), vision–language feature fusion (UGLF \cite{tran2024unifying}), and enhanced spatial–temporal feature injection combined with contrastive learning (ASTRM \cite{santra2025precise}). Multimodal approaches such as ASTRA \cite{xarles2023astra} further extend PES by incorporating audio cues to complement visual information. Notably, all these methods rely on fully supervised learning, requiring extensive frame-level annotations.

\subsection{Unsupervised Learning}
Unsupervised learning, particularly in the form of self-supervised learning, has emerged as an effective paradigm for pretraining deep models without relying on manual annotations. By designing surrogate learning objectives, self-supervised methods enable models to learn meaningful representations that can be transferred to downstream tasks with limited labeled data.

Among these approaches, DINO \cite{caron2021emerging} has demonstrated that strong visual representations can be learned through self-distillation by enforcing consistency between different augmented views of the same image, without using any labels. Models pretrained with DINO achieve competitive performance after fine-tuning on small labeled datasets. Building upon this idea, Teeti et al. \cite{Teeti_2023_ICCV} extends self-distillation to videos by introducing asymmetric temporal contexts, where the teacher observes a broader temporal window to encourage the student to predict future information and capture temporal dependencies.

Alternatively, VideoMAE \cite{tong2022videomae} adopts a masked autoencoding strategy with a high masking ratio, training models to reconstruct masked video content and thereby learn compact spatiotemporal representations. While these methods have shown strong performance in representation learning, they are primarily designed for pretraining and do not directly address task-specific supervision such as frame-level event localization.

\subsection{Semi-Supervised Learning}
Semi-supervised learning aims to leverage large amounts of unlabeled data alongside a limited set of labeled samples \cite{chen2022semi}, often incorporating techniques from self-supervised learning. SSL has been extensively studied in 2D vision tasks such as image classification, image retrieval, and object detection. Motivated by these successes, recent works have increasingly explored SSL in the video domain \cite{chen2022semi}.
Most image or video-based SSL methods are built upon two core principles: consistency regularization and pseudo-labeling \cite{van2020survey}. Consistency regularization encourages a model to produce invariant predictions under perturbations in the input or model space, while pseudo-labeling assigns high-confidence predictions on unlabeled data as supervisory signals during training.
Consistency-based approaches typically rely on data augmentations or multi-view sampling to enforce feature or prediction consistency across different views of the same video. Representative methods include FixMatch \cite{sohn2020fixmatch}, SVFormer \cite{xing2023svformer}, End-to-End Semi-Supervised Learning for Video \cite{kumar2022end}, VideoSSL \cite{jing2021videossl}, and recent extensions tailored for video representation learning \cite{kwon2025boosting}.
In contrast, pseudo-labeling-based approaches focus on generating reliable labels for unlabeled data, often through teacher–student frameworks or multi-view agreement mechanisms \cite{vandeghen2022semi, nag2022semi, singh2024semi, xiong2021multiview, xia2023learning}.
Recent work by Liu et al.~\cite{liu2025few} represents one of 
The first attempts to apply SSL to sports video event spotting 
via multimodal knowledge distillation, where a student RGB 
model is guided by a teacher, integrating pose estimation, ball 
tracking, and court detection. While effective, this reliance 
on multiple domain-specific pretrained models limits 
scalability and generalisability. In contrast, our approach 
operates purely on RGB frames without any auxiliary components.

\section{Method}
\subsection{Problem Definition}
We consider PES in sports videos under a semi-supervised setting. Given an input clip \(X \in \mathbb{R}^{T \times H \times W \times 3}\), the task is to predict an event class for each frame. The training set \(\mathcal{D} = \{X_i\}_{i=1}^{N}\) is divided into a labeled subset \(\mathcal{D}_l\), which provides frame-level event annotations, and an unlabeled subset \(\mathcal{D}_u\), which contains no annotations. Our objective is to improve frame-level event classification by exploiting \(\mathcal{D}_u\) together with the limited labeled data in \(\mathcal{D}_l\).

\begin{figure}
    \centering
    \includegraphics[width=1\linewidth]{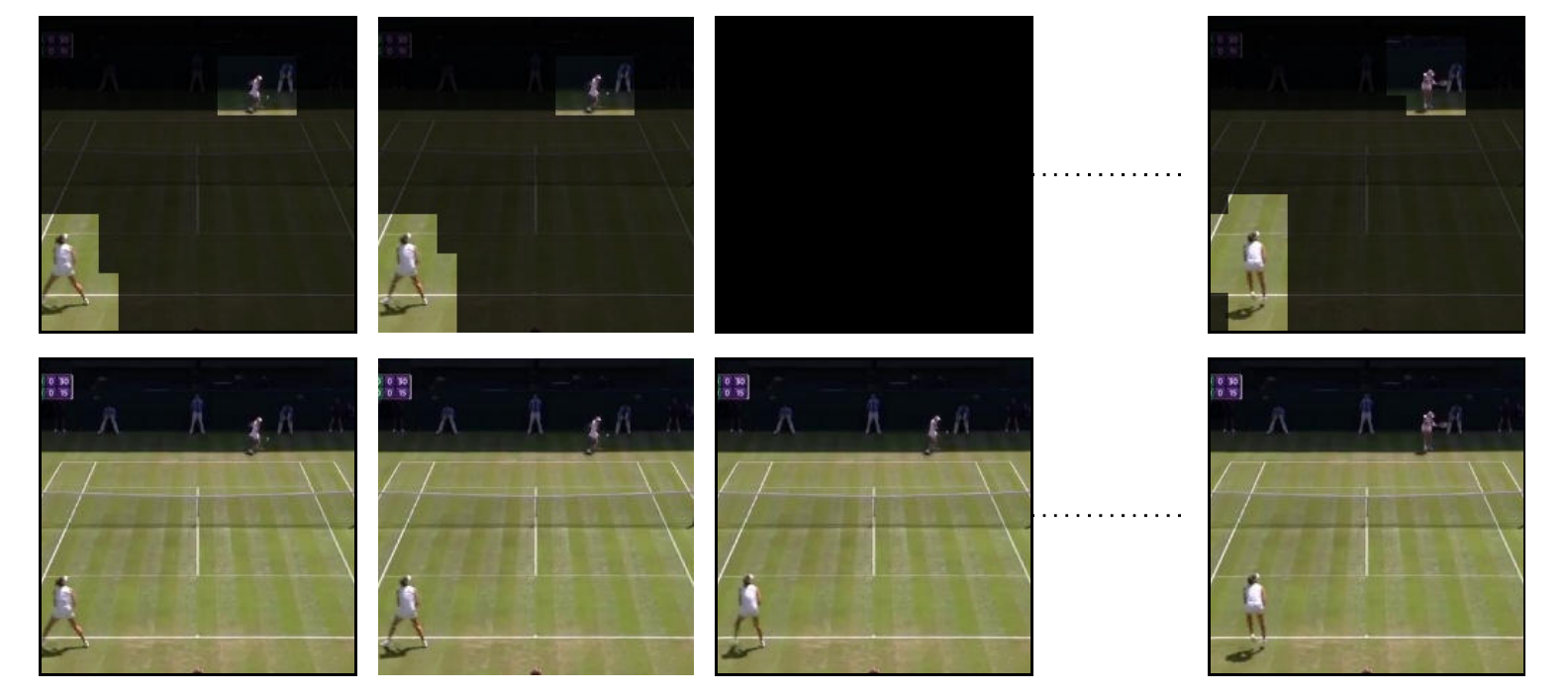}
    \caption{Effect of the proposed Temporal Motion Augmentation. The top row shows augmented frames, while the bottom row presents the corresponding original images.}
    \label{fig:augmentation}
\end{figure}

\subsection{Transformer Gate Shift (TGS)}
Vision Transformers (ViTs) have been widely adopted across computer vision tasks, including self-supervised representation learning in DINO~\cite{caron2021emerging}, due to their strong ability to capture fine-grained spatial details and the interpretability offered by self-attention mechanisms. In contrast, existing fully supervised approaches for PES predominantly rely on 2D CNN backbones for their simplicity and efficiency. When combined with temporal modules such as MFS~\cite{xu2025multifocustemporalshiftingprecise}, GSM~\cite{sudhakaran2020gate}, GSF~\cite{sudhakaran2023gate}, and TSM~\cite{lin2019tsm}, CNN-based models can capture temporal information without introducing dedicated video backbones. However, these temporal modules are primarily designed for convolutional architectures, and effective temporal modelling mechanisms for 2D ViTs remain underexplored.

To address this limitation, we adopt a Vision Transformer backbone and propose TGS as shown in Figure \ref{fig:overview}, a temporal module tailored for ViTs. TGS enables temporal information exchange while preserving the core token-processing structure of image-based transformers. In practice, TGS is inserted at the end of each ViT encoder block, allowing temporal information to be progressively injected throughout the backbone.
Let the input token features be $\mathbf{X} \in \mathbb{R}^{B \times T \times N \times C}$, where $B$ denotes the batch size, $T$ the temporal length, $N$ the number of tokens per frame, and $C$ the feature dimension. We first apply layer normalization and ReLU activation to $\mathbf{X}$, and then reshape the features to $\tilde{\mathbf{X}} \in \mathbb{R}^{(B \cdot N) \times C \times T}$ so that temporal convolution can be applied independently to each token sequence along the temporal dimension.
To capture temporal context at multiple scales, we apply depthwise 1D convolutions with kernel sizes $k_d = 2d + 1$, where $d \in \mathcal{D} = \{1,3,5\}$. For each scale, the temporally aggregated feature is passed through an MLP that preserves the channel dimension $C$, followed by a sigmoid activation to generate a full element-wise gate:
\begin{equation}
\mathbf{G}^{(d)} = \sigma \left( \mathrm{MLP}\left( \mathrm{Conv1D}^{(d)}(\tilde{\mathbf{X}}) \right) \right),
\end{equation}
where $\mathrm{Conv1D}^{(d)}$ operates along the temporal dimension and $\mathbf{G}^{(d)} \in \mathbb{R}^{(B \cdot N) \times C \times T}$ is the resulting gate tensor. Unlike GSM, GSF, and related gating-based temporal modules that typically generate coarse gates for only a few channel groups, TGS produces a gate value for every feature element. This full element-wise gating allows finer control over temporal information flow and helps preserve subtle local cues that are important for precise event spotting.
The input features are split equally along the channel dimension into three groups,
\begin{equation}
\mathbf{X} = \left[\mathbf{X}_{\text{past}},\; \mathbf{X}_{\text{future}},\; \mathbf{X}_{\text{static}}\right],
\end{equation}
and the corresponding gates are partitioned in the same way:
\begin{equation}
\mathbf{G}^{(d)} = \left[\mathbf{G}^{(d)}_{\text{past}},\; \mathbf{G}^{(d)}_{\text{future}},\; \mathbf{G}^{(d)}_{\text{static}}\right].
\end{equation}

Each component is modulated by its corresponding gate:
\begin{equation}
\tilde{\mathbf{X}}^{(d)}_{\ast} =
\mathbf{G}^{(d)}_{\ast} \odot \mathbf{X}_{\ast},
\quad
\ast \in \{\text{past}, \text{future}, \text{static}\},
\end{equation}
where $\odot$ denotes element-wise multiplication. Only the gated past and future groups are temporally shifted, while the static group remains unchanged. This design enables feature-wise control over whether and how temporal information is exchanged across frames.

Temporal shifting is then applied as
\begin{equation}
\mathbf{Y}^{(d)} =
\mathrm{Concat}\Big(
\mathrm{Shift}_{-d}(\tilde{\mathbf{X}}^{(d)}_{\text{past}}),\;
\mathrm{Shift}_{+d}(\tilde{\mathbf{X}}^{(d)}_{\text{future}}),\;
\tilde{\mathbf{X}}^{(d)}_{\text{static}}
\Big),
\end{equation}
where $\mathrm{Shift}_{-d}$ and $\mathrm{Shift}_{+d}$ denote temporal shifts toward the past and future, respectively.

The outputs from different temporal scales are fused using learnable softmax-normalized weights:
\begin{equation}
\mathbf{Y} = \sum_{d \in \mathcal{D}} \alpha_d \mathbf{Y}^{(d)}, \qquad
\alpha_d = \frac{\exp(w_d)}{\sum_{j \in \mathcal{D}} \exp(w_j)}.
\end{equation}

Finally, a projection layer and a residual connection are applied:
\begin{equation}
\mathbf{Z} = \mathbf{X} + \mathrm{Proj}(\mathbf{Y}),
\end{equation}
yielding temporally enhanced token representations that can be seamlessly integrated into standard Vision Transformer blocks.

\subsection{Temporal Motion Augmentation (TMA)}
Consistency regularization typically trains a student to match a teacher under stronger augmentations, encouraging invariance to input perturbations. While effective in image-based learning, directly applying standard image augmentations to sports videos is suboptimal. Random cropping or masking may remove small but critical cues such as the ball, and large static background regions can dominate the augmented views, reducing sensitivity to event-relevant motion.

To address this, we propose TMA, shown in Figure~\ref{fig:augmentation}, which is designed to emphasize motion-sensitive regions in sports videos. Our design is motivated by the observation that the most informative elements in PES are typically moving objects, such as the ball and player contact regions.

Concretely, we compute a motion map using frame differences between adjacent frames, which highlights moving regions without requiring external components such as tracking or detection models. The motion map is divided into non-overlapping spatial patches, and each patch is assigned a motion score by averaging its top $2\%$ highest-motion pixels. This top-percentile strategy helps preserve small but fast-moving objects that would otherwise be overwhelmed by static background areas. Patches are then ranked by motion score, and the highest-motion regions are preserved while the remaining background regions are down-weighted, thereby encouraging the model to focus on motion-relevant cues.

In addition, we introduce temporal masking by randomly dropping multiple short temporal segments from each clip. This forces the model to infer missing information from surrounding frames and encourages stronger temporal reasoning. Together, these spatial and temporal augmentations produce more informative views for consistency-based learning in semi-supervised PES.

\begin{figure*}
    \centering
    \includegraphics[width=0.8\linewidth]{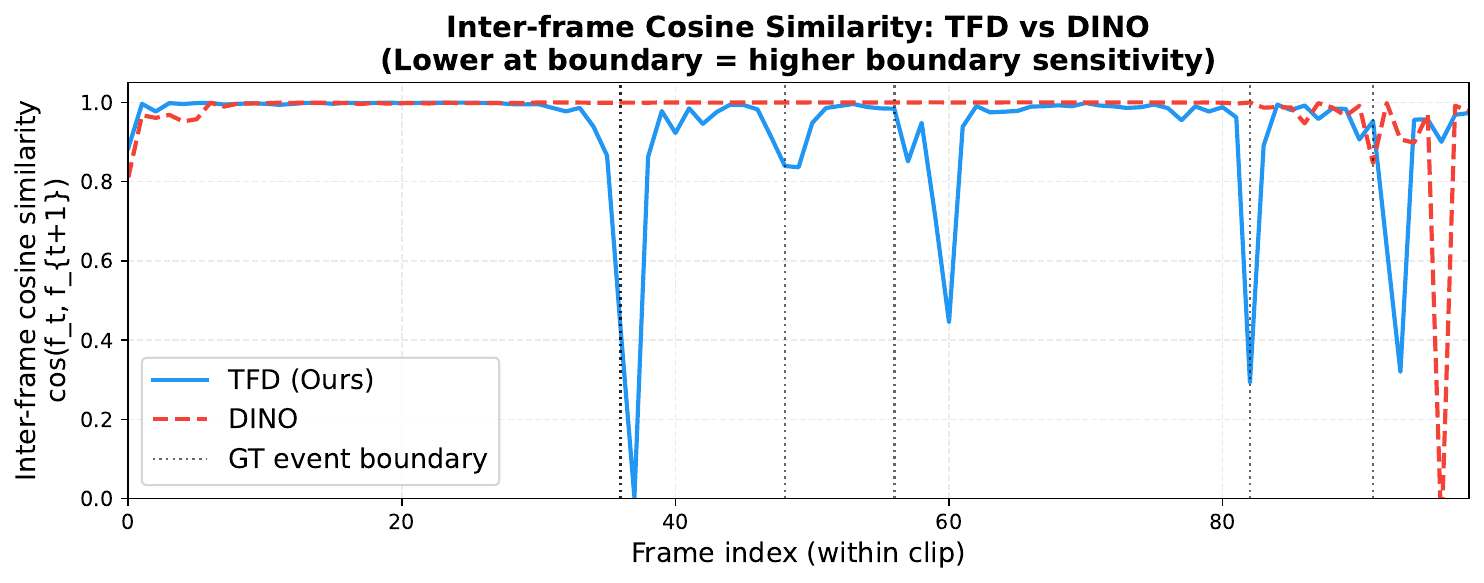}
    \caption{Inter-frame cosine similarity between adjacent backbone features for TFD (ours) and DINO on a representative Tennis clip. TFD features consistently exhibit sharp similarity drops at ground-truth event boundaries (dotted lines), indicating high boundary sensitivity. DINO features remain near-constant (\~1.0) across most boundaries, confirming that projection-level invariance suppresses the temporal discriminability required for precise event spotting.}
    \label{fig:cosine_sim}
\end{figure*}

\subsection{Temporal Feature Distillation (TFD)}
\label{sec:tfd}
While TGS already yields strong performance under fully supervised training, effectively leveraging large amounts of unlabeled sports videos remains essential for further improving generalization. To this end, we adopt a self-distillation framework inspired by DINO, while introducing task-driven modifications specifically tailored for PES.

Given an input video clip
\(\mathbf{X} \in \mathbb{R}^{B \times T \times C \times H \times W}\),
we generate two augmented views using weak and strong augmentation strategies, denoted as $\mathbf{X}^w$ and $\mathbf{X}^s$, respectively. The student network processes the strongly augmented view $\mathbf{X}^s$, while the teacher network processes the weakly augmented view $\mathbf{X}^w$. The teacher parameters are updated as an exponential moving average (EMA)~\cite{tarvainen2017mean} of the student parameters, providing stable targets during training.

Unlike standard DINO-style self-distillation, which aligns projection-head 
outputs to promote semantic invariance, such invariance is less suitable for 
PES. PES requires distinguishing between temporally adjacent frames that are 
visually similar yet semantically distinct. Enforcing projection-level 
invariance can suppress subtle temporal boundaries and motion-sensitive cues 
that are critical for accurate event localization. In sports videos, this may 
bias the model toward dominant large-scale motion, such as player movement, 
while reducing sensitivity to small but semantically important objects such as 
the ball, as illustrated in Figure~\ref{fig:attn_visualization}.
This limitation can be partially understood through DINO’s prototype-based design. DINO learns a large set of prototypes to capture diverse semantic categories in image datasets. However, frames within a short sports video clip typically occupy a much narrower semantic space and exhibit high visual similarity. As a result, projection-head alignment may encourage neighboring frames to map to a limited subset of prototypes, reducing temporal separability and weakening sensitivity to fine-grained event transitions. This reveals a mismatch between projection-level invariance and the frame-level localization objective of PES.

To empirically validate this claim, Figure~\ref{fig:cosine_sim} visualises 
the inter-frame cosine similarity $\cos(f_t, f_{t+1})$ between adjacent 
backbone features for both TFD and DINO on a representative Tennis clip. 
DINO features remain near-constant (${\geq}0.999$) across almost all 
ground-truth event boundaries, confirming that projection-level invariance 
suppresses temporal discriminability. In contrast, TFD features exhibit sharp 
similarity drops at boundaries --- reducing to as low as $0.003$ at one 
boundary compared to $0.999$ for DINO --- while maintaining a non-boundary 
average of $0.936$, demonstrating that TFD preserves boundary-sensitive 
representations without collapsing overall feature quality.

To address this issue, we introduce Temporal Feature Distillation, which aligns temporally structured backbone features produced by the TGS-enhanced ViT directly, rather than projected representations. By distilling features that explicitly encode multi-scale temporal dependencies, TFD better preserves motion-sensitive and boundary-aware information that is crucial for frame-level event spotting.
In addition, we observe that purely unsupervised pretraining is insufficient for PES. For computational efficiency, sports videos are typically resized to relatively low resolutions, where small but critical objects such as tennis or table tennis balls may occupy only a few pixels. Without supervised guidance, these subtle motion cues can be treated as noise and ignored by the model. To mitigate this issue, we adopt a supervised warm-up strategy, in which the model is first trained on labeled data to establish event-aware and object-sensitive representations. The unlabeled distillation loss is then gradually introduced using a ramp-up schedule, enabling the model to exploit unlabeled data without suppressing small yet essential event cues. Together, these design choices enable effective semi-supervised learning for PES by combining the stability of self-distillation with task-aware temporal representations and motion-focused supervision.

\paragraph{Distillation Objective.}
Let the teacher and student networks extract backbone features from the weakly and strongly augmented views:
\begin{equation}
\mathbf{F}^t = \mathcal{T}(\mathbf{X}^w), \quad 
\mathbf{F}^s = \mathcal{S}(\mathbf{X}^s),
\qquad \mathbf{F}^t, \mathbf{F}^s \in \mathbb{R}^{B \times T \times D},
\end{equation}
where $\mathcal{T}$ denotes the EMA teacher and $\mathcal{S}$ the student. Features are $\ell_2$-normalized along the channel dimension, and gradients are stopped on the teacher branch.

\textbf{Temporal alignment loss.}
We align temporally corresponding frames by minimizing cosine distance between teacher and student features:
\begin{equation}
\mathcal{L}_{\mathrm{align}}
=
\frac{1}{B|\Omega|}
\sum_{b=1}^{B}\sum_{t \in \Omega}
\left(
1 - 
\cos\!\left(
\mathbf{F}^{s}_{b,t},\ 
\mathrm{sg}(\mathbf{F}^{t}_{b,t})
\right)
\right),
\end{equation}
where $\Omega=\{1,\ldots,T\}$ denotes the set of temporal indices (identical for the weak and strong views), and $\mathrm{sg}(\cdot)$ indicates the stop-gradient operator.

\textbf{Supervised loss.}
For labeled data, the event spotting head is trained with the standard cross-entropy loss:
\begin{equation}
\mathcal{L}_{\mathrm{sup}} = \mathrm{CE}(\hat{\mathbf{y}}, \mathbf{y}),
\end{equation}
where $\hat{\mathbf{y}}$ denotes the predicted class probabilities and $\mathbf{y}$ the ground-truth event labels.

\textbf{Overall objective.}
The final training objective combines the supervised loss and the temporal feature distillation loss:
\begin{equation}
\mathcal{L}_{\mathrm{TFD}}
=
\mathcal{L}_{\mathrm{sup}}
+
\lambda(t)\,\mathcal{L}_{\mathrm{align}},
\end{equation}
The weight $\lambda(t)$ is introduced after a supervised warm-up stage and is progressively increased using a cosine ramp-up schedule. Specifically, the model is first trained using only $\mathcal{L}_{\mathrm{sup}}$ for 5 epochs. The distillation term is then activated and its weight is gradually increased from 0 to 1 over the next 10 epochs, after which it remains fixed at 1 for the rest of training.

\begin{table*}[t]
\centering
\caption{
Performance comparison (mAP \%) under different annotation ratios (10\%, 20\%, 40\%, 80\%) across fine-grained sports datasets. Our method demonstrates clear advantages in low-label regimes while remaining competitive as supervision increases. Complete results for all annotation ratios are provided in the supplementary material.}
\label{tab:label-ratio-results}
\renewcommand{\arraystretch}{1.0}
\small
\begin{adjustbox}{width=\textwidth}
\begin{tabular}{l |cccc |cccc |cccc |cccc}
\toprule
\textbf{Model} &
\multicolumn{4}{c}{\textbf{Tennis}} &
\multicolumn{4}{c}{\textbf{FS\_Comp}} &
\multicolumn{4}{c}{\textbf{FS\_Perf}} &
\multicolumn{4}{c}{\textbf{FineDiving}} \\
& 
\textbf{0.1} & \textbf{0.2} & \textbf{0.4} & \textbf{0.8} &
\textbf{0.1} & \textbf{0.2} & \textbf{0.4} & \textbf{0.8} &
\textbf{0.1} & \textbf{0.2} & \textbf{0.4} & \textbf{0.8} &
\textbf{0.1} & \textbf{0.2} & \textbf{0.4} & \textbf{0.8} \\
\midrule

\multicolumn{17}{l}{\textbf{Fully Supervised}} \\
\midrule
ASTRM~\cite{santra2025precise} &
76.61& \textbf{86.33} & 90.46 & 92.97&
31.71& 55.42 & 65.94 & 72.49 &
45.75& 61.69& 74.14& 79.96&
45.50& 57.02&  62.86&  66.19\\

E2E-200~\cite{hong2022spotting} &
64.38& 79.70 & 88.11 &  91.84&
23.72& 51.01 & 63.66& 72.67 &
37.20& 59.72& 73.08& 79.14&
39.12& 55.08& 56.88& 63.35 \\


T-Deed~\cite{xarles2024t}&
36.69 & 65.04 & 80.21& 91.44&
15.21& 41.93& 56.96 & 67.43 &
20.32&  51.81& 68.40 & 77.95 &
34.31& 51.12& 58.06& 65.89\\

MFS \cite{xu2025multifocustemporalshiftingprecise}& 
37.84& 71.15& 86.39& 91.38& 
16.37& 49.24& 63.33& 70.08& 
 29.34& 57.55& 73.39& 78.45& 
40.77& 55.02& 60.58& 65.39\\

ViT-Tiny \cite{dosovitskiy2020image}&
10.96 & 44.60& 60.40& 71.30 &
 2.65& 11.55&  23.57& 47.99 &
2.13& 17.27& 39.74 & 57.71 &
34.65& 48.42& 57.87&  64.32  \\

ViT-Small \cite{dosovitskiy2020image}&
46.32& 60.61& 73.42& 80.34&
25.23& 35.33& 53.53 & 65.58 &
 27.80& 38.71& 40.03 & 59.34 &
43.68& 53.38& 60.61 & 63.88 \\

\midrule
\multicolumn{17}{l}{\textbf{SSL \& USL}} \\
\midrule

Kwon et al.~\cite{kwon2025boosting} &
70.13& 81.33 & 88.42 & 89.55&
17.81& 52.29& 66.34& 66.95&
36.61 & 65.35& 74.91 & 75.93 &
52.01 & 58.28& 62.90 & \textbf{67.92}\\

Teeti et al. ~\cite{Teeti_2023_ICCV} & 
72.08 & 83.61& 89.27& 92.10&
24.08& 48.95& 63.18&  75.82&
45.07& 62.49 & 68.10& 82.71&
51.92 & 58.33&  61.75& 67.66\\

Zhou et al. \cite{zhou2024towards}& 
79.76& 82.86& 91.92& 92.39& 
37.94 & 56.80& 60.30& 74.73 & 
42.62&  61.03  &  51.67 & 80.85 & 
50.15 & 59.08& 61.10& 65.02\\

Dino \cite{caron2021emerging}&
 69.55 & 75.62& 83.92& 88.92&
24.00& 34.08& 57.10& 63.04  &
 18.82& 42.62& 67.67& 72.28 &
41.82& 45.62& 48.98&55.70\\

\midrule
\multicolumn{17}{l}{\textbf{Ours}} \\
\midrule


\textbf{\textit{Ours (ViT-S)}} &
\textbf{80.19} & 85.70 &  \textbf{90.81}& \textbf{93.04}  &
 \textbf{38.29} & \textbf{65.58}  &  \textbf{70.20}&  \textbf{79.27}&
\textbf{50.29}& \textbf{69.08} & \textbf{77.63}& \textbf{84.52}&
\textbf{52.87}& \textbf{60.33}  & \textbf{63.22} & 66.57\\

\bottomrule
\end{tabular}
\end{adjustbox}
\end{table*}

\section{Datasets}

To evaluate the effectiveness of our method, we conduct experiments on several widely adopted PES benchmarks, including \textbf{Tennis}, \textbf{Figure Skating}, and \textbf{FineDiving}.
\textbf{Tennis} \cite{hong2022spotting} contains 3,345 video clips collected from 28 professional tennis matches, recorded at 25 or 30 FPS. The dataset includes 33,791 frame-level event annotations across six event categories. Following the standard protocol, videos from 19 matches are used for training and validation, while the remaining 9 matches are reserved for testing.
\textbf{Figure Skating} \cite{hong2021video} includes 371 short-program performances extracted from 11 broadcast videos recorded at 25 FPS during Winter Olympics and World Championships events. The dataset provides 3,674 annotated events across four classes. Following \cite{hong2022spotting}, we evaluate on two standard splits: the Competition Split (FSComp) and the Performance Split (FSPerf).
\textbf{FineDiving} \cite{xu2022finediving} contains 3,000 diving clips with temporal annotations, totaling 7,010 events across four classes. We convert the annotations into frame-level labels following the evaluation protocol in \cite{hong2022spotting} to ensure consistency with existing PES benchmarks.

\section{Evaluations}
\subsection{Experimental Details}
All experiments are conducted on 8 NVIDIA A100 GPUs. Unless otherwise specified, models are trained for 50 epochs with a batch size of 4 using the AdamW optimizer~\cite{loshchilov2017decoupled}. The learning rate and weight decay are both set to $1\times10^{-4}$, and the default clip length is 100 frames.
In semi-supervised training, the teacher model is updated as an EMA of the student following Mean Teacher~\cite{tarvainen2017mean}. We use a base EMA decay of 0.9995 and a fast EMA decay of 0.99. We first perform a supervised warm-up for 5 epochs. The unlabeled distillation loss is then introduced progressively using a cosine ramp-up schedule over 10 epochs, with the maximum weighting coefficient $\lambda$ set to 1. 
For TMA, each frame is partitioned into a $16 \times 16$ grid. We identify the 9 grid cells with the largest temporal variation and apply the augmentation with probability 0.5. For frame masking, the mask length is set to 3, with at most 5 masked segments per video clip. More details can be found in the supplementary materials.

\subsection{Evaluation Protocol}
We evaluate our method under semi-supervised settings across all datasets by varying the proportion of labeled training data. Specifically, we consider 10\%, 20\%, 40\%, and 80\% labeled data, covering both low-label and label-rich regimes, with particular emphasis on the low-label setting. For fair comparison, all methods are trained with three different random seeds, and the reported results are averaged across runs.
We follow the standard evaluation protocols adopted in recent state-of-the-art PES methods~\cite{hong2022spotting,santra2025precise,xarles2024t}. Performance is measured using Average Precision within a temporal tolerance of $\delta$ frames (AP@$ \delta $). AP is computed independently for each event class, and mean Average Precision (mAP) is obtained by averaging across all classes. Following prior work, we emphasize strict localization performance, particularly at $\delta = 1$. Results under other tolerance settings, including $\delta \in \{0,2\}$, are provided in the supplementary materials.

\begin{table}[t]
\centering
\caption{Ablation on Tennis and FSPerf with ViT backbones. Results are mAP (\%) at $\delta=1$ using 10\% labeled data. (1) evaluates the contributions of TGS and TFD across backbone scales; (2) compares TFD with different view generation strategies, where $\Delta$ is relative to \textit{TFD + Standard Views}; and (3) studies clip length for the full model, where $\Delta$ is relative to the default setting $L=100$.}
\label{tab:ablation_main}
\renewcommand{\arraystretch}{1.15}
\small
\resizebox{1\linewidth}{!}{
\begin{tabular}{llcccc}
\toprule
\multirow{2}{*}{\textbf{Section}} & \multirow{2}{*}{\textbf{Variant}} & 
\multicolumn{2}{c}{\textbf{Tennis}} & 
\multicolumn{2}{c}{\textbf{FSPerf}} \\
\cmidrule(lr){3-4}\cmidrule(lr){5-6}
& & mAP & $\Delta$ & mAP & $\Delta$ \\
\midrule

\multicolumn{6}{l}{\textit{(1) Architectural Contribution}} \\
& ViT-Tiny                     & 10.96& --     & 2.13  & --     \\
& ViT-Tiny + TGS               & 44.77 & +33.81& 9.70  & +7.57\\
& ViT-Tiny + TFD               & 52.60& +41.64& 8.61& +6.48\\
& ViT-Tiny + TGS + TFD         & 72.26 & +61.30& 24.09 & +21.96\\
& ViT-Small                    & 51.56 & --     & 28.61 & --     \\
& ViT-Small + TGS              & 68.39 & +16.83 & 41.37 & +12.76 \\
& ViT-Small + TFD              & 60.85& +9.29& 29.50& +0.89\\
& ViT-Small + TGS + TFD        & 80.19 & +11.80 & 50.29 & +8.92  \\
\midrule

\multicolumn{6}{l}{\textit{(2) TFD Under Different View Generation Strategies}} \\
& TFD + Standard Views         & 75.32 & --     & 48.56 & --     \\
& TFD + Temporal Aug. (Ours)   & 80.19 & +4.87  & 50.29 & +1.73 \\
\midrule

\multicolumn{6}{l}{\textit{(3) Varying Clip Length (Full Model)}} \\
& $L=50$   & 73.50 & -6.69 & 55.85 & +5.56 \\
& $L=75$   & 77.38 & -2.81 & 56.97 & +6.68 \\
& $L=100$  & 80.19 & --    & 50.29 & --    \\
& $L=125$  & 76.72 & -3.47 & 58.97 & +8.68 \\
& $L=200$  & 76.04 & -4.15 & 58.78 & +8.49 \\
\bottomrule
\end{tabular}}
\end{table}

\textbf{Baselines.} We compare our method against recent fully supervised, semi-supervised, and self-supervised approaches. For methods not originally designed for PES, particularly SSL and USL methods, we adapt them to the PES setting under a unified implementation protocol. In particular, all ViT-based baselines use the same ViT-S + TGS backbone, together with the same input resolution, clip length, optimizer, training schedule, and evaluation protocol unless method-specific components require otherwise. This controlled setting reduces confounding factors from backbone architecture and temporal modeling capacity, allowing the comparison to focus mainly on differences in learning objectives. Since some SSL/USL methods were originally proposed with different architectural or procedural assumptions, these results should be interpreted as controlled adaptations to PES rather than exact reproductions of each method’s original optimal setting. For fully supervised baselines, results under 100\% annotation are provided in the supplementary material.

\subsection{Performance Comparison}
Table~\ref{tab:label-ratio-results} presents quantitative comparisons with recent fully supervised, SSL, and USL approaches under varying annotation ratios.
Overall, our method achieves the best mAP in \textbf{14 out of 16} evaluation settings, demonstrating consistent superiority across datasets and supervision regimes. The advantage is particularly pronounced in low-label scenarios, where effective utilization of unlabeled data becomes critical.
Under the 10\% annotation setting, our method consistently delivers the largest performance gains. For example, on FSPerf with 10\% labeled data, the strongest competing method (ASTRM) achieves 45.75 mAP, whereas our method reaches 50.29 mAP, yielding a substantial improvement of \textbf{+4.54}. Similar trends are observed across all four datasets, where our approach outperforms both fully supervised and SSL baselines under limited supervision.
As the amount of labeled data increases, our method remains competitive and maintains leading performance in most cases. On Tennis, where even 10\% supervision already provides strong learning signals, our approach still achieves the best results at higher ratios (e.g., 93.04 mAP at 80\%), demonstrating that the gains are not limited to low-resource settings.
In summary, the results indicate that our method (1) substantially improves performance in low-label regimes, (2) scales effectively as supervision increases, and (3) generalizes well across datasets of varying difficulty.

\subsection{Ablations}
We conduct ablation studies to analyze the contribution of each design component, as shown in Table~\ref{tab:ablation_main}. 
All results are reported under a strict temporal tolerance of $\delta = 1$, and experiments are performed on two representative datasets (Tennis and FSPerf) to demonstrate that the observed trends are consistent across different sports scenarios. 
All ablations are conducted under the 10\% labeled setting with identical training configurations unless otherwise specified.

\textbf{TGS.}
We first evaluate the impact of the proposed TGS under fully supervised training by comparing both ViT-Tiny and ViT-Small backbones with and without TGS. As shown in Table~\ref{tab:ablation_main}, adding TGS consistently improves performance across both model scales. For ViT-Tiny, TGS boosts mAP from 10.96 to 44.77 on Tennis and from 2.13 to 9.70 on FSPerf. For ViT-Small, it further improves performance from 51.56 to 68.39 on Tennis and from 28.61 to 41.37 on FSPerf, corresponding to gains of \textbf{+16.83 mAP} and \textbf{+12.76 mAP}, respectively. These results show that TGS effectively injects temporal modeling capacity into ViT backbones, with consistent benefits across different model sizes.

\textbf{TFD.}
We further analyze the proposed TFD objective under different view generation strategies. Even with standard augmentation, TFD already outperforms the fully supervised ViT baseline, showing its ability to exploit unlabeled data effectively. However, TFD without TGS is suboptimal, as vanilla ViT primarily captures spatial features and lacks explicit temporal modeling. When combined with TGS, TFD becomes substantially more effective because the teacher--student alignment is performed on features that already encode meaningful temporal patterns. 

\begin{figure}[t]
    \centering
    \includegraphics[width=0.9\linewidth]{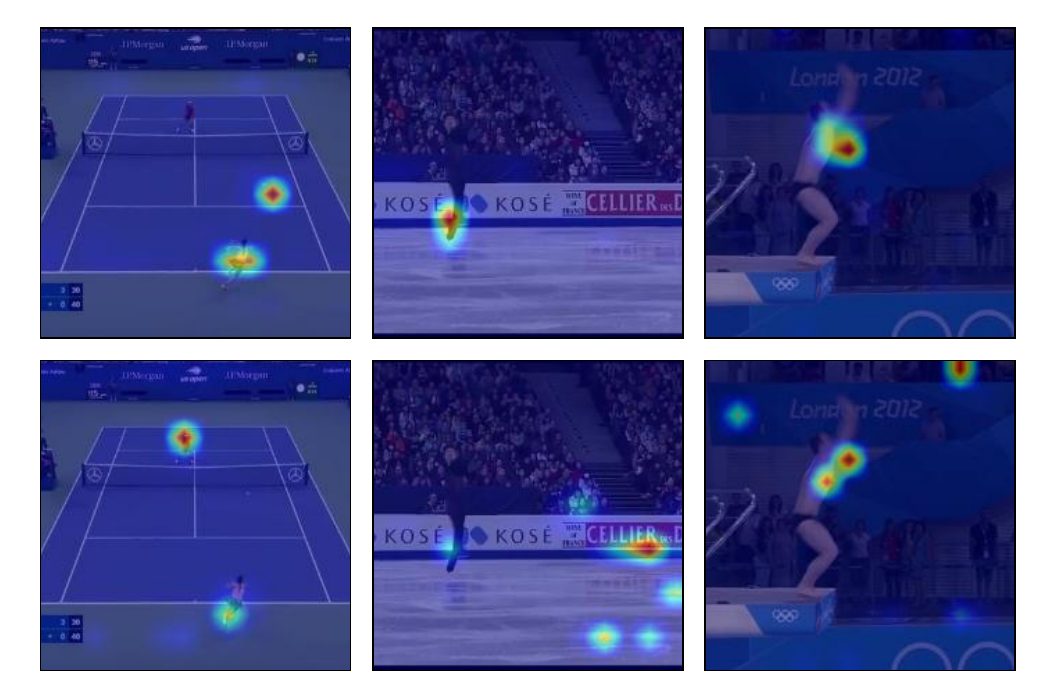}
    \caption{Attention map comparison between our TFD (top) and 
    vanilla DINO (bottom) across Tennis, Figure Skating, and Diving. 
    Our model attends to event-relevant regions such as the ball, while DINO diffuses attention across 
    dominant background structures.}
    \label{fig:attn_visualization}
\end{figure}

Figure~\ref{fig:attn_visualization} further supports this observation through attention map comparisons with vanilla DINO across three sports. DINO often focuses mainly on the player while neglecting small but important cues such as the ball, or produces diffuse attention over the whole frame. In contrast, our method consistently attends to event-relevant regions, including the ball and player contact areas. This more localized and discriminative attention pattern is consistent with the quantitative gains in Table~\ref{tab:label-ratio-results}, especially in low-label regimes.

\textbf{TMA.}
TMA emphasizes motion-salient regions when constructing paired temporal views before applying TFD. By encouraging feature alignment to focus on dynamically changing regions rather than static background content, the model learns more discriminative and event-relevant representations. As shown in Table~\ref{tab:ablation_main}, this leads to a clear improvement (+4.87 mAP) on Tennis, reaching \textbf{80.19 mAP}. The gain is more pronounced on Tennis likely because the camera is usually static, making motion regions easier to separate from the background. In contrast, the improvement on FS is more limited, as camera motion often co-occurs with athlete motion, making motion cues less distinctive. These results confirm the importance of motion-aware temporal pairing for effective feature distillation in PES.

\textbf{Clip Length}. We further study the effect of clip length on performance. As shown in Table~\ref{tab:ablation_main}, the optimal temporal context is dataset-dependent. On Tennis, performance degrades consistently as clip length deviates from the default $L=100$, with drops of up to $-6.69$ mAP at $L=50$, suggesting that Tennis events are best captured within a moderate and precise temporal window. In contrast, on FSPerf, all alternative clip lengths outperform the default setting, with both shorter clips ($L=75$, $+6.68$) and longer clips ($L=125$, $+8.68$; $L=200$, $+8.49$) yielding consistent gains. This indicates that FSPerf benefits from a different temporal context, where shorter clips may reduce irrelevant motion while longer clips better capture the preparation and follow-through phases of figure skating elements. Nevertheless, we adopt $L=100$ as the default setting because it is a standard choice in prior PES methods \cite{hong2021video, santra2025precise} and provides a practical balance between temporal coverage, computational cost, and overall performance across datasets. These results suggest that while $L=100$ serves as a strong and consistent default, clip length remains an important dataset-specific hyperparameter in sports video understanding.

\textbf{Model Complexity.}
Table~\ref{tab:complexity} reports parameter count, GFLOPs, and FPS. Parameter count is computed as the total number of learnable parameters. GFLOPs are obtained from the measured MACs using the standard conversion $1\text{ MAC}=2\text{ FLOPs}$, and FPS is measured as the number of processed frames per second during inference on a single clip of 100 frames after warm-up. Since CNN-based and ViT-based PES models differ substantially in their underlying computational characteristics, direct efficiency comparisons across these architectural families are not entirely fair. We therefore focus on the incremental overhead introduced by TGS relative to the vanilla ViT-S backbone. Compared with ViT-S, TGS increases the parameter count by 5.44M (from 23.45M to 28.89M), the computational cost by 355.85 GFLOPs (from 644.30 to 1000.15), and reduces inference speed from 1649 FPS to 867 FPS. Although this introduces a non-trivial overhead, it consistently improves spotting performance. Under fully supervised training, TGS improves mAP by \textbf{+16.83} on Tennis and \textbf{+12.76} on FSPerf over ViT-S, with larger gains observed when combined with TFD and temporal augmentation. These results indicate that TGS provides a favorable accuracy--complexity trade-off within the ViT-based PES setting.

\begin{table}[t]
\centering
\caption{Model complexity comparison. CNN-based and ViT-based methods are grouped separately because their computational characteristics are not directly comparable. FPS is measured under the same inference setting for all models.}
\label{tab:complexity}
\small
\begin{tabular}{lccc}
\toprule
Model & Params (M) & GFLOPs & FPS \\
\midrule
\multicolumn{4}{l}{\textit{CNN-based}} \\
E2E-200            & 4.45  & 42.19  & 5392 \\
MFS                & 4.46  & 42.92  & 3053 \\
ASTRM              & 8.69  & 134.47 & 4177 \\
T-DEED             & 16.42 & 43.35  & 3995 \\
\midrule
\multicolumn{4}{l}{\textit{ViT-based}} \\
ViT-Tiny              &  5.97 & 182.63  & 4497\\
ViT-Tiny + TGS              & 7.37 & 273.41  & 2184 \\
ViT-Small              & 23.45 & 644.30  & 1649 \\
ViT-Small + TGS  & 28.89 & 1000.15 & 867  \\
\bottomrule
\end{tabular}
\end{table}

\section{Conclusion}
In this paper, we propose a semi-supervised framework for PES that improves both temporal modeling and unlabeled data utilization. We introduce TGS to inject temporal motion cues into ViT backbones, TFD to align temporally meaningful features across augmented views, and TMA to emphasize motion-salient regions during semi-supervised training. Experiments on multiple fine-grained sports datasets show consistent improvements across annotation ratios, with especially strong gains in low-label settings. These results highlight the importance of task-aware temporal representation learning for fine-grained sports video understanding. More detailed discussions on limitations and future directions are provided in the supplementary material.

\clearpage
\bibliographystyle{ACM-Reference-Format}
\bibliography{sample-base}

\end{document}